\documentclass[10pt,twocolumn,letterpaper]{article}

\usepackage{iccv}
\usepackage{times}
\usepackage{epsfig}
\usepackage{graphicx}
\usepackage{amsmath}
\usepackage{amssymb}
\usepackage{multirow}
\usepackage{subfigure}
\usepackage{booktabs}

\usepackage[pagebackref=true,breaklinks=true,letterpaper=true,colorlinks,bookmarks=false]{hyperref}
\usepackage{cleveref}

\iccvfinalcopy 

\begin{document}

\title{TIFace: Improving Facial Reconstruction through Tensorial Radiance \\Fields and Implicit Surfaces}

\author{Ruijie Zhu, Jiahao Chang, Ziyang Song, Jiahuan Yu, Tianzhu Zhang\\
University of Science and Technology of China\\
{\tt\small \{ruijiezhu, cjhdemail, songziyang, yjhcs1998\}@mail.ustc.edu.cn, tzzhang@ustc.edu.cn}
}

\maketitle











\noindent{\textbf{Abstract. }} 
This report describes the solution that secured the first place in the "View Synthesis Challenge for Human Heads (VSCHH)" at the ICCV 2023 workshop.
Given the sparse view images of human heads, the objective of this challenge is to synthesize images from novel viewpoints.
Due to the complexity of textures on the face and the impact of lighting, the baseline method TensoRF yields results with significant artifacts, seriously affecting facial reconstruction.
To address this issue, we propose TI-Face, which improves facial reconstruction through tensorial radiance fields (T-Face) and implicit surfaces (I-Face), respectively. 
Specifically, we employ an SAM-based approach to obtain the foreground mask, thereby filtering out intense lighting in the background. 
Additionally, we design mask-based constraints and sparsity constraints to eliminate rendering artifacts effectively.
The experimental results demonstrate the effectiveness of the proposed improvements and superior performance of our method on face reconstruction. 
The code will be available at~\url{https://github.com/RuijieZhu94/TI-Face}.\\

\section{Introduction}
Reconstructing 3D objects from multi-view 2D images is a fundamental and long-standing challenge in the fields of computer vision and graphics, with critical applications in robotics, 3D modeling, virtual reality, and beyond. 
Traditional methods~\cite{schoenberger2016sfm, schoenberger2016mvs} typically involve finding matching point pairs across multiple viewpoints and leveraging the principles of multi-view geometry to reconstruct 3D objects. 
However, these approaches face difficulties in handling scenes with a lack of texture or repetitive patterns and often struggle to generate dense reconstructions.

Therefore, some learning-based multi-view stereo (MVS) methods have been proposed~\cite{yao2018mvsnet,gu2020cascade,unimvsnet}, using plane assumptions to aggregate cost volumes and predict dense depth end-to-end for the recovery of 3D objects. 
However, they struggle to effectively handle occlusions and non-Lambertian surfaces. 
Recently, NeRF~\cite{mildenhall2021nerf} and a plethora of subsequent works~\cite{wang2021neus,Chen2022tensorf,muller2022instant} have been introduced, demonstrating the significant power of neural networks in the implicit representation of 3D objects. 
By learning to map 3D coordinates to volume density and view-dependent colors, these methods efficiently achieve novel view synthesis and produce remarkably realistic rendering effects. 
However, they are often optimized for a specific scene, making the training of the network lack prior knowledge.

\begin{figure}
\centering
\subfigure[Data collection]{
    \includegraphics[width=0.45\linewidth]{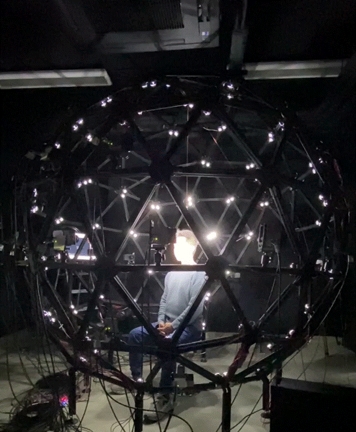}
}
\subfigure[Captured images]{
    \includegraphics[width=0.45\linewidth]{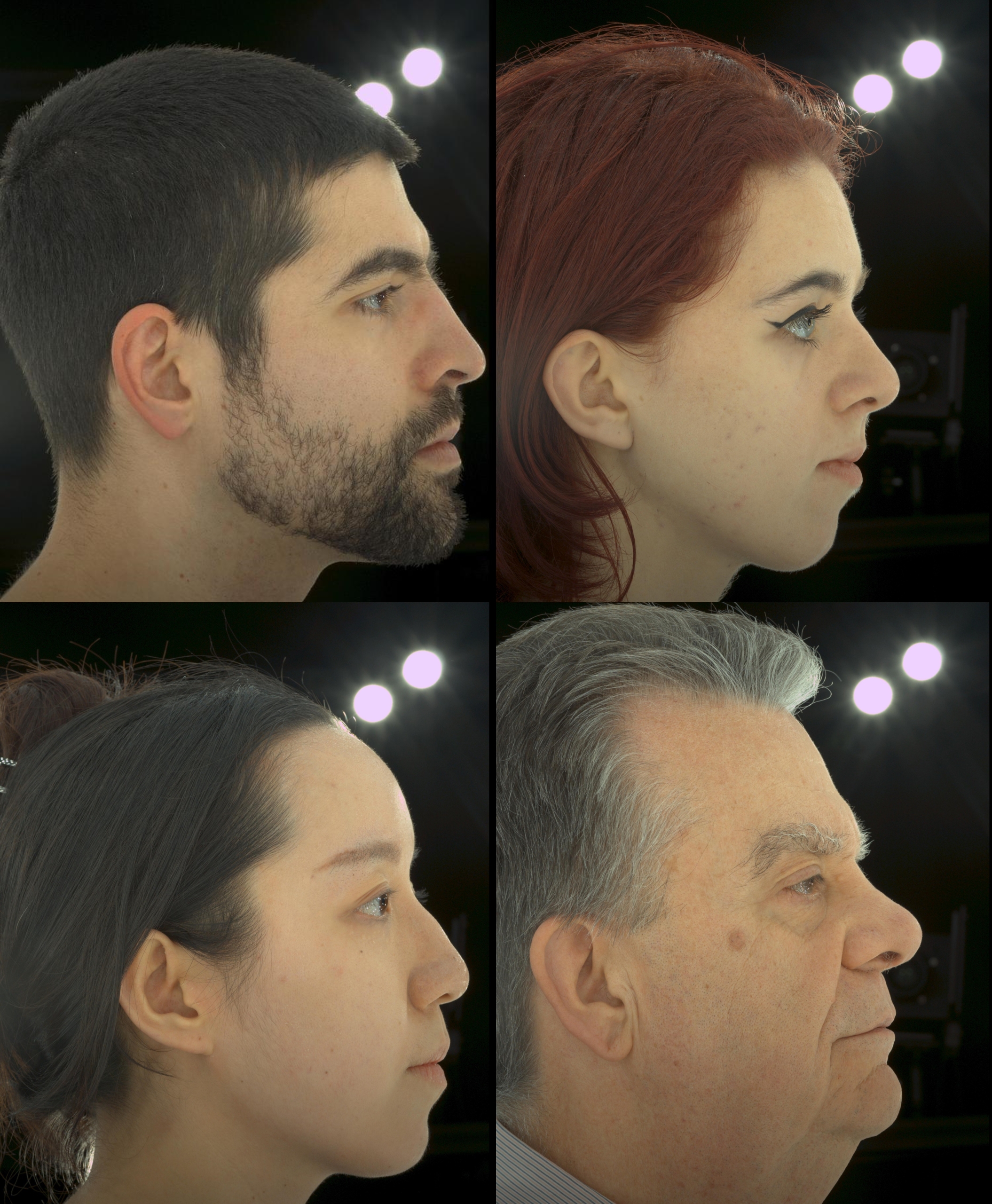}
}
\caption{Illustration of the ILSH dataset~\cite{Zheng_2023_ilsh}. The images are captured from a light stage environment. For the same human head, images from multiple viewpoints are captured simultaneously to ensure geometric consistency across multiple views.}
\end{figure}

To explore efficient and realistic representations of 3D objects, a novel challenge~\cite{Jang_2023_vschh} was introduced at the ICCV 2023 workshop\footnote{\url{https://sites.google.com/view/vschh/home}}. 
Various methods are evaluated in a new public dataset~\cite{Zheng_2023_ilsh} for their performance in the task of synthesizing new views of human heads. 
Given a set of sparse views of the human head, this challenge requires the reconstruction of images from specified new perspectives. 
As views are limited and sparse, and include uncertain halos as background noise, existing methods for novel view synthesis face significant challenges in this context.

To address this challenge, we propose TI-Face, which enhances facial reconstruction through tensorial radiance fields and implicit surfaces.
Firstly, we analyze the shortcomings of the baseline method TensoRF~\cite{Chen2022tensorf}, specifically the artifacts caused by background noise. 
To address this issue, we employ an SAM-based image segmentation algorithm, ViTMatte~\cite{yao2023vitmatte}, which achieves high-precision segmentation of the background region with manual placement of a few prompt points. 
The proposed TI-Face further constrains the volume density sampling points in the background region based on the generated mask, ensuring that the color of the background region is transparent rather than black.
On the other hand, we introduce I-Face, which reconstructs human heads by enhancing implicit surface reconstruction through the combination of InstantNGP~\cite{muller2022instant} and NeuS~\cite{wang2021neus}. 
By combining mask constraints and sparsity constraints, I-Face can rapidly and efficiently reconstruct the surface of human heads, rendering surface colors from new perspectives.
By combining T-Face and I-Face, our approach faithfully reconstructs realistic human heads, particularly preserving fine details in the facial features. 
The proposed method surpassed the approaches of other participants in VCSHH, securing the first place and highlighting the outstanding performance of TI-Face in the task of novel view synthesis.

The main contributions of our work are as follows:
\begin{itemize}
\item We propose a novel framework TI-Face, which improves facial reconstruction through tensorial radiance fields (T-Face) and implicit surfaces (I-Face).
\item To remove floating artifacts during rendering, we adopt an SAM-based mask generation method and make mask constraints on both T-Face and I-Face. Meanwhile, a sparsity loss is additionally applied to I-Face for better rendering quality.
\item Experiments on the ILSH dataset demonstrate the effectiveness of the proposed improvements. Furthermore, we win the first place in VSCHH at the ICCV 2023 workshop.
\end{itemize}

\section{Methodology}

\subsection{Baseline}
To address the View Synthesis Challenge for Human Heads (VSCHH), we use TensoRF~\cite{Chen2022tensorf} as our baseline.
TensoRF utilizes 4D tensors to model the radiance field of a scene. The key idea mainly focuses on the low-rank factorization of 4D tensors to achieve better rendering quality and smaller model size with fast speed.
Following the baseline model, we implement a vector-matrix decomposition version of TensoRF with some modifications to fit the ILSH dataset.
As shown in~\Cref{fig:baseline}, although the baseline model achieves relatively accurate facial reconstruction, the reconstructed results often have floating artifacts, which severely affects the reconstruction quality of the edges of the face.
We speculate that this is because there are light sources in the background area of the rendered image, which causes a sudden change in the background color.
To avoid this, we use an efficient way to get the masks corresponding to the input images. 
Furthermore, we improve both explicit and implicit rendering methods and build additional constraints based on masks to improve rendering quality.

\subsection{T-Face}
\begin{figure}[h]
\begin{center}
   \includegraphics[width=\linewidth]{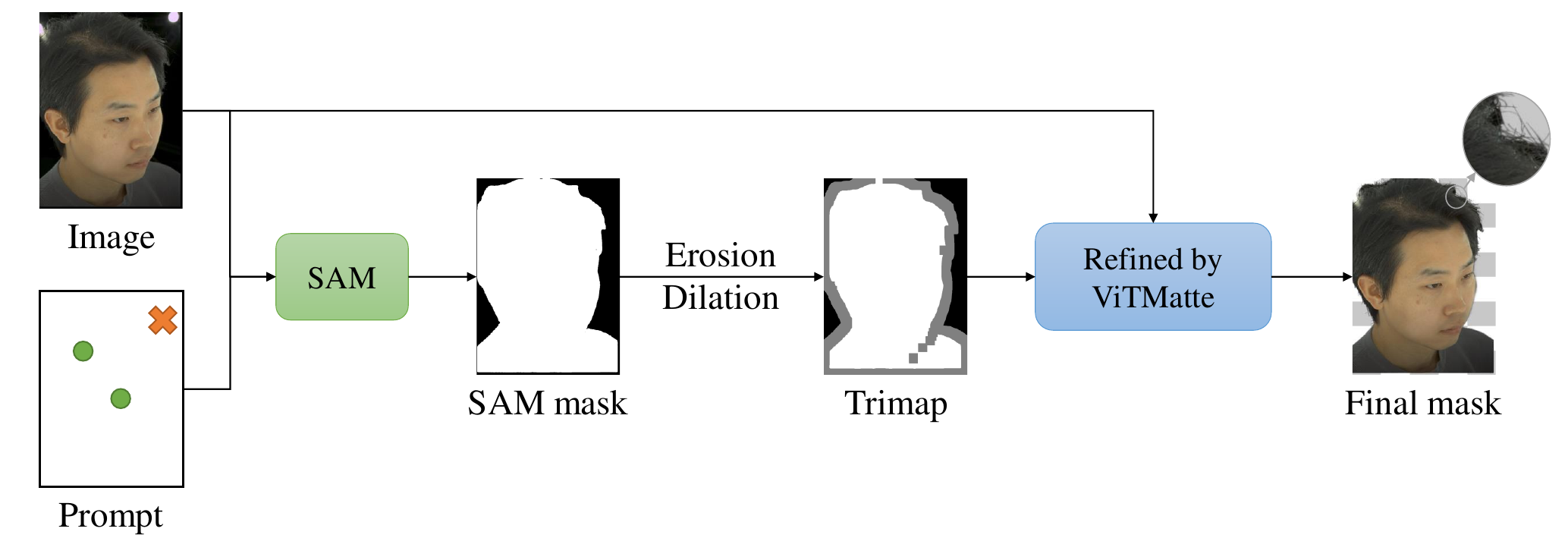}
\end{center}
   \caption{The pipeline of mask generation. To reduce the computational overhead, we first downsample the image and then upsample the predicted mask. Then we label the image with prompt points and use SAM to obtain a initial mask. After erosion and dilation on the image, we generate a trimap mask, where the gray areas represent the areas that need further segmentation. With the help of VitMatte~\cite{yao2023vitmatte}, we obtain a final refined mask.}
\label{fig:vitmatte}
\end{figure}

Recently, SAM-based image segmentation methods have attracted a lot of attention~\cite{kirillov2023segment}. 
Following ViTMatte~\cite{yao2023vitmatte}, we obtain the corresponding masks of the input images through a small number of label points as prompts.
As shown in~\Cref{fig:vitmatte}, we get masks fine enough to distinguish the foregrounds and the backgrounds of the images, even in the hair strand region.

\begin{figure}[h]
\begin{center}
   \includegraphics[width=0.9\linewidth]{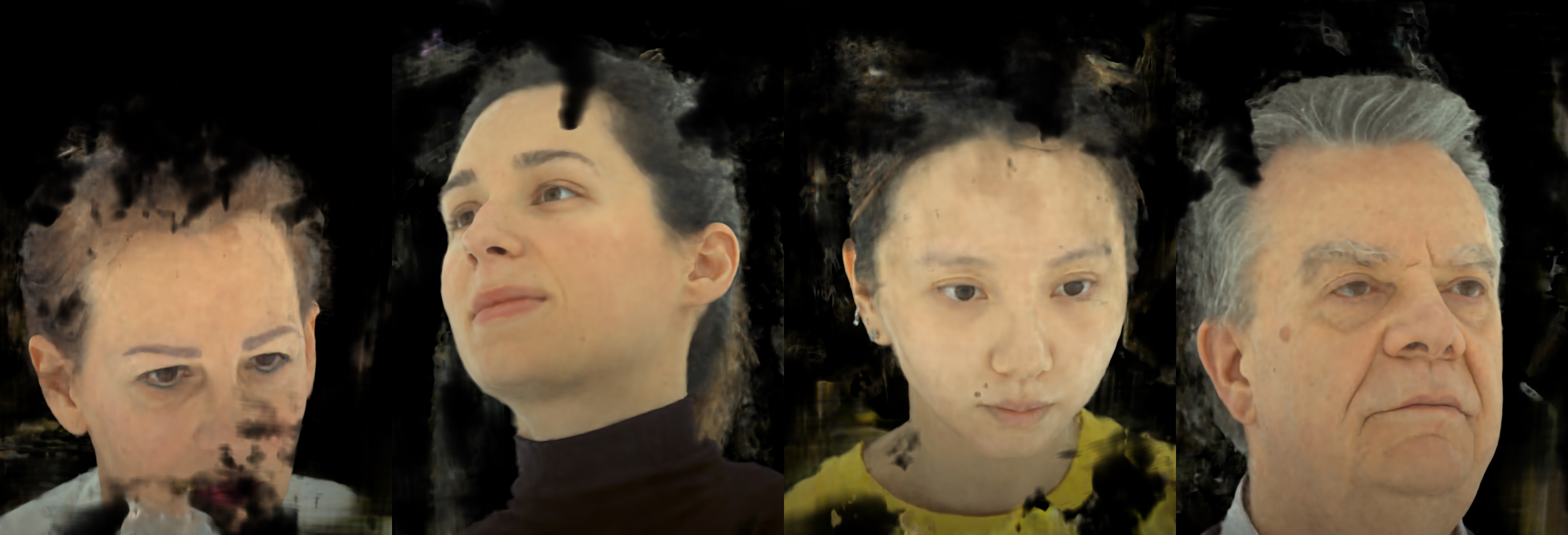}
\end{center}
   \caption{The failure cases of baseline model based on masks. }
\label{fig:failmask}
\end{figure}

A natural way to utilize masks is to combine masks with RGB images into RGBA images, as implemented on the Blender dataset.
This actually uses masks to set the values of background area in the image to a fixed color (e.g. black).
Although it appears to remove the effect of cluttered backgrounds on rendering faces, this method still exhibits some artifacts in our experiments, as shown in~\Cref{fig:mask} and~\Cref{fig:failmask}.
We speculate that the reason is that the generated mask is not completely accurate and that the model lacks constraints on the background, resulting in a not so fine sampling on the surface.

To address this issue, we propose a constraint on the mask and demonstrate its effectiveness through experiments.
For each pixel, we march along a ray, sampling $Q$ shading points along the ray and computing the accumulated density weights:
\begin{equation}
    T = \sum_{q=1}^{Q} \tau_q (1-\exp(-\sigma_q \Delta_q )),\tau_q =\exp(-\sum_{p=1}^{q-1}\sigma_q \Delta_q ).
\end{equation}
Here, $\sigma_q$ is the density at the sample location $x_q$, $\Delta_q$ is the ray step size, and $\tau_q$ denotes the ray transmittance.
The mask loss is:
\begin{equation}
    \mathcal{L}_{mask}=\frac{\lambda }{|B|}\sum_{x}T_x^2\mathbb{I}(x\in B),
\end{equation}
where $x$ is the pixel corresponding to the ray, $B$ is the background areas according to the masks, $\mathbb{I}(\cdot)$ is the indicator function. Here, we set $\lambda = 0.01$ in our experiments.
Note that we also tried to use cross entropy in mask loss function, and found that it is not as effective as the proposed loss.

Finally, the total loss is defined as:
\begin{equation}
    \mathcal{L}_{T} = \mathcal{L}_{color} + \alpha \mathcal{L}_{reg} + \beta \mathcal{L}_{mask} + \gamma \mathcal{L}_{sparsity},
\end{equation}
where $\mathcal{L}_{color}$ is a MSE loss on RGB color, $\mathcal{L}_{reg}$ is the regularization term and $\mathcal{L}_{sparsity}$ is the total variation (TV) loss that measures the difference between neighboring values in the matrix or vector factors.

\subsection{I-Face}
To further improve the face rendering quality, we explore another way to render photo-realistic faces. 
We observe that although T-Face shows promising results, it does not perform well in facial reconstruction details.
Inspired by InstantNGP~\cite{muller2022instant} and NeuS~\cite{wang2021neus}, we use implicit surface rendering for face reconstruction based on Instant-NeuS implementation~\cite{instant-nsr-pl}.
Before this, we also tried Instant-NeRF~\cite{instant-nsr-pl} (a combination of InstantNGP~\cite{muller2022instant} and NeRF~\cite{mildenhall2021nerf}) to complete face reconstruction, but gave up due to the difficulty of imposing constraints on neural radiance fields to remove artifacts.
In the experiment, we use the binary cross entropy loss as the mask loss, as mentioned in NeuS~\cite{wang2021neus}.
To avoid floating artifacts, we also add sparsity loss:
\begin{equation}
    \mathcal{L}_{sparsity} = \frac{1}{|S|} \sum_{y\in S} \exp (-\gamma d(y)),
\end{equation}
where $y$ is the sampled point, $d(y)$ is the corresponding SDF values, 
$S$ represents the collection of sampling points.
Here we set $\gamma = 0.5$.
Finally, the total loss is defined as:
\begin{equation}
    \mathcal{L}_{I} = \mathcal{L}_{color} + \alpha \mathcal{L}_{reg} + \beta \mathcal{L}_{mask} + \gamma \mathcal{L}_{sparsity},
\end{equation}
where $\mathcal{L}_{color}$ is a MSE loss on RGB color and $\mathcal{L}_{reg}$ is the Eikonal regularization~\cite{gropp2020implicit}.

\subsection{Model Ensemble}
To further improve the results, we use a simple and effective linear weighted ensemble method to obtain the final results.
In practice, we use a set of linear weights $(0.1, 0.6, 0.3)$ corresponding to the baseline, T-Face, and I-Face to group the rendered images.

\section{Experiments}
This section presents the experimental results during the challenge phase. 
We first compare our approach with baseline method and then analyse the effectiveness of the proposed improvements.

\subsection{Implementation Details}
We implement our T-Face and I-Face in PyTorch. 
In T-Face, we follow the baseline configurations, using Adam optimizer~\cite{kingma2014adam} with $(\beta_1, \beta_2)=(0.9, 0.99)$ and initial learning rate of $0.02$ for tensor factors and $0.001$ for the MLP decoder.
We train our model T-Face for 50000 steps with a batch size of 4096 rays on a single NVIDIA RTX 3090 (40-60 minutes per scene). 
Following TensoRF~\cite{Chen2022tensorf}, we start from an initial coarse grid with $128^3$ voxels and then upsample them at steps 2000, 3000, 4000, 5500, and 7000 to a fine grid with $300^3$ voxels.
In I-Face, we use AdamW optimizer~\cite{loshchilov2017decoupled} with $(\beta_1, \beta_2)=(0.9, 0.99)$ and the learning rate $0.01$. 
We train our model I-Face for 20000 steps with a dynamic batch size (256-8192) on a single NVIDIA RTX 3090 (10-20 minutes per scene).

\begin{figure}[h]
\begin{center}
   \includegraphics[width=0.9\linewidth]{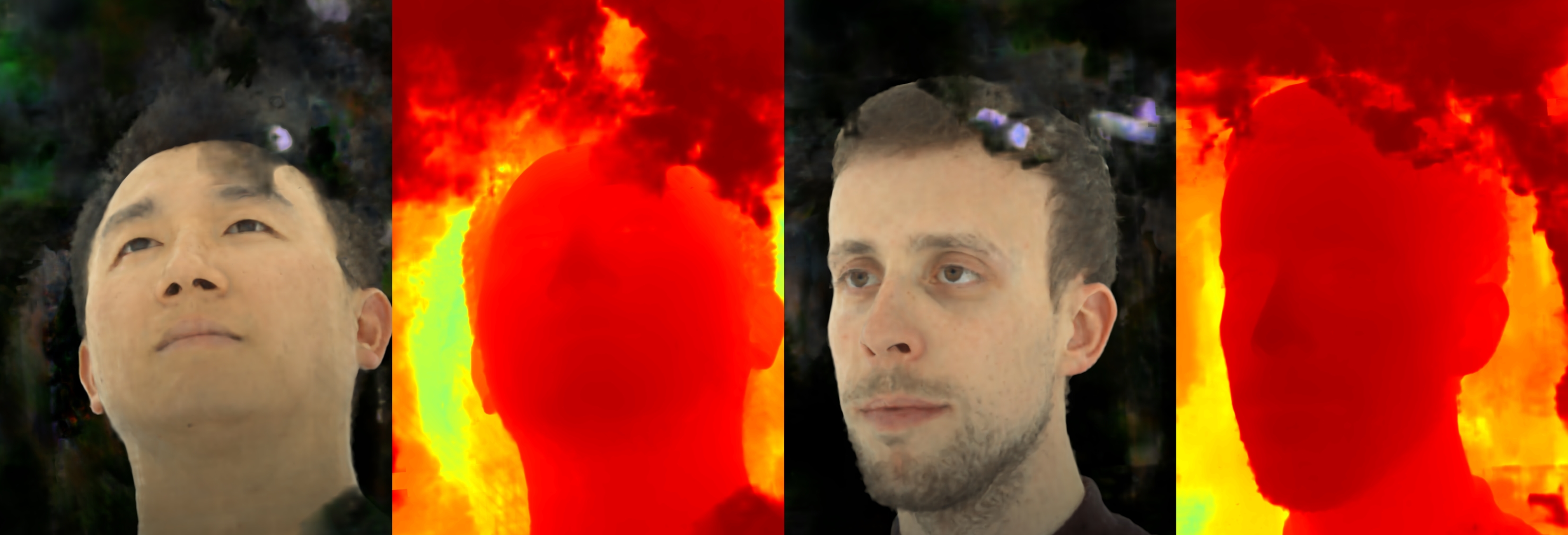}
\end{center}
   \caption{The rendering results of baseline model. Left: rendered image. Right: depth image.}
\label{fig:baseline}
\end{figure}

\begin{figure}[h]
\begin{center}
   \includegraphics[width=0.9\linewidth]{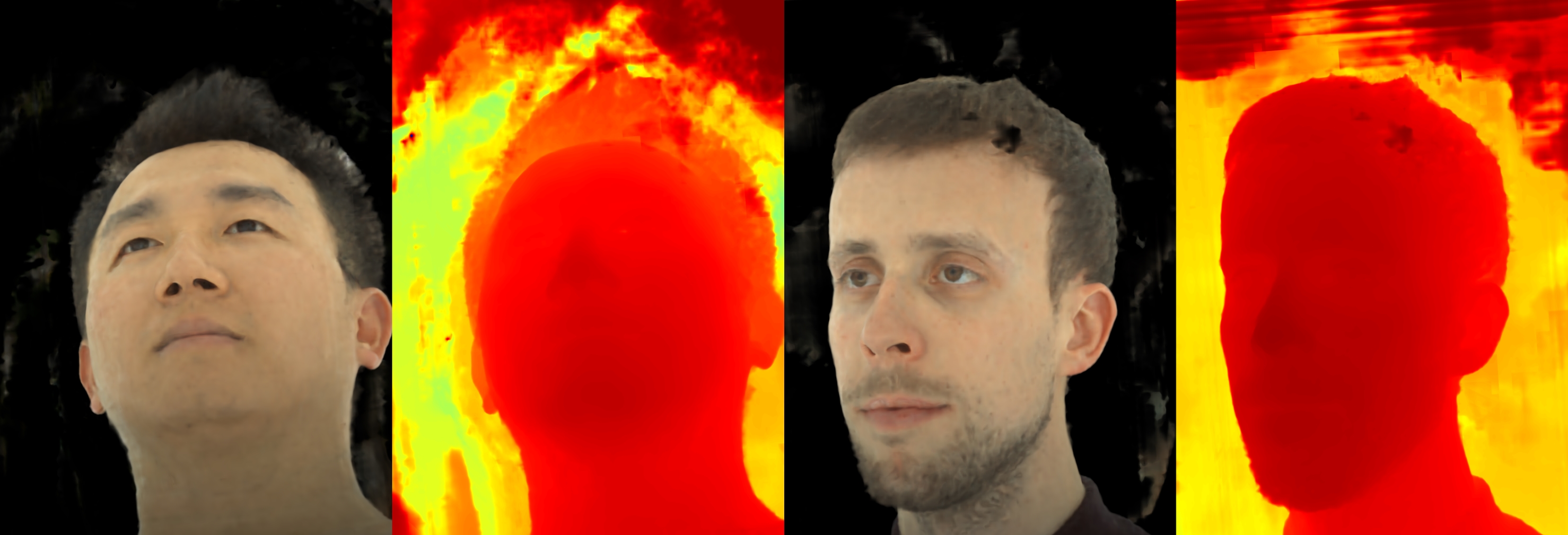}
\end{center}
   \caption{The rendering results of baseline model based on masks. Left: rendered image. Right: depth image.}
\label{fig:mask}
\end{figure}

\begin{table*}
\begin{center}
\begin{tabular}{ccccccc}
\hline
\multirow{2}{*}{Methods} & \multicolumn{2}{c}{Full Region} & \multicolumn{2}{c}{Masked Region} & \multirow{2}{*}{ToI (Sec.)} & \multirow{2}{*}{Devices} \\
\cmidrule(lr){2-3} \cmidrule(lr){4-5}
 & PSNR$\uparrow$ & SSIM$\uparrow$ & \textbf{PSNR}$\uparrow$ & SSIM$\uparrow$ & & \\
\hline
DINER-SR* & 22.37 & 0.72 & 28.50 & \textbf{0.83} & 87.25 & V100 \\
MPFER-H* & \textbf{28.05} & \textbf{0.84} & \textbf{28.90} & \textbf{0.83} & \textbf{1.50} & V100 \\
\hline
KHAG & \textbf{22.14} & 0.64 & 23.39 & 0.79 & \textbf{2.58} & RTX A6000\\
xoft & 20.01 & 0.64& 25.02 & 0.80 & 727.00 & A10\\
Y-KIST-NeRF & 20.73 & \textbf{0.71} & 25.54 & 0.82 & 15.10 & RTX 6000 ADA\\
CUBE & 21.07 & 0.66 & 25.72 & 0.81 & 95.00 & A100 \& H100\\
CogCoVi & 21.49 & 0.70 & 26.33 & 0.82 & 806.00 & A40 \\
NoNeRF & 20.37 & 0.69 & 26.43 & 0.82 & 175.58 & RTX3090 \\
TI-Face(Ours) & 21.66 & 0.68 & \textbf{27.02} & \textbf{0.83} & 76.88 & RTX 3090\\
\hline
\end{tabular}
\end{center}
\caption{Evaluation Results of methods at the VSCHH leaderboard~\cite{Jang_2023_vschh}. Methods with * are submitted by the challenge organizers.}
\label{table: compare}
\end{table*}

\subsection{Experimental Results}

\noindent{\textbf{Comparison with other methods.}}
We compare our TI-Face with other methods sumbitted to the VSCHH benchmark~\cite{Jang_2023_vschh}. 
The VSCHH selects the PSNR metric of mask region to be the ranking indicator, which emphasize the rendering quality of high-value areas (\ie, face and hair).
Masks for high-value areas are generated by the organizers and are invisible to participants in the challenge.
As show in~\Cref{table: compare}, the experimental results demonstrate that our TI-Face significantly outperforms methods from other challenge participants in the evaluation of masked image region.
Meanwhile, compared with the second and third place methods NoNeRF and CogCoVi, our method TI-Face has a clear advantage in time of inference (ToI), reflecting the high efficiency of the proposed method.

\begin{figure}[h]
\begin{center}
   \includegraphics[width=0.98\linewidth]{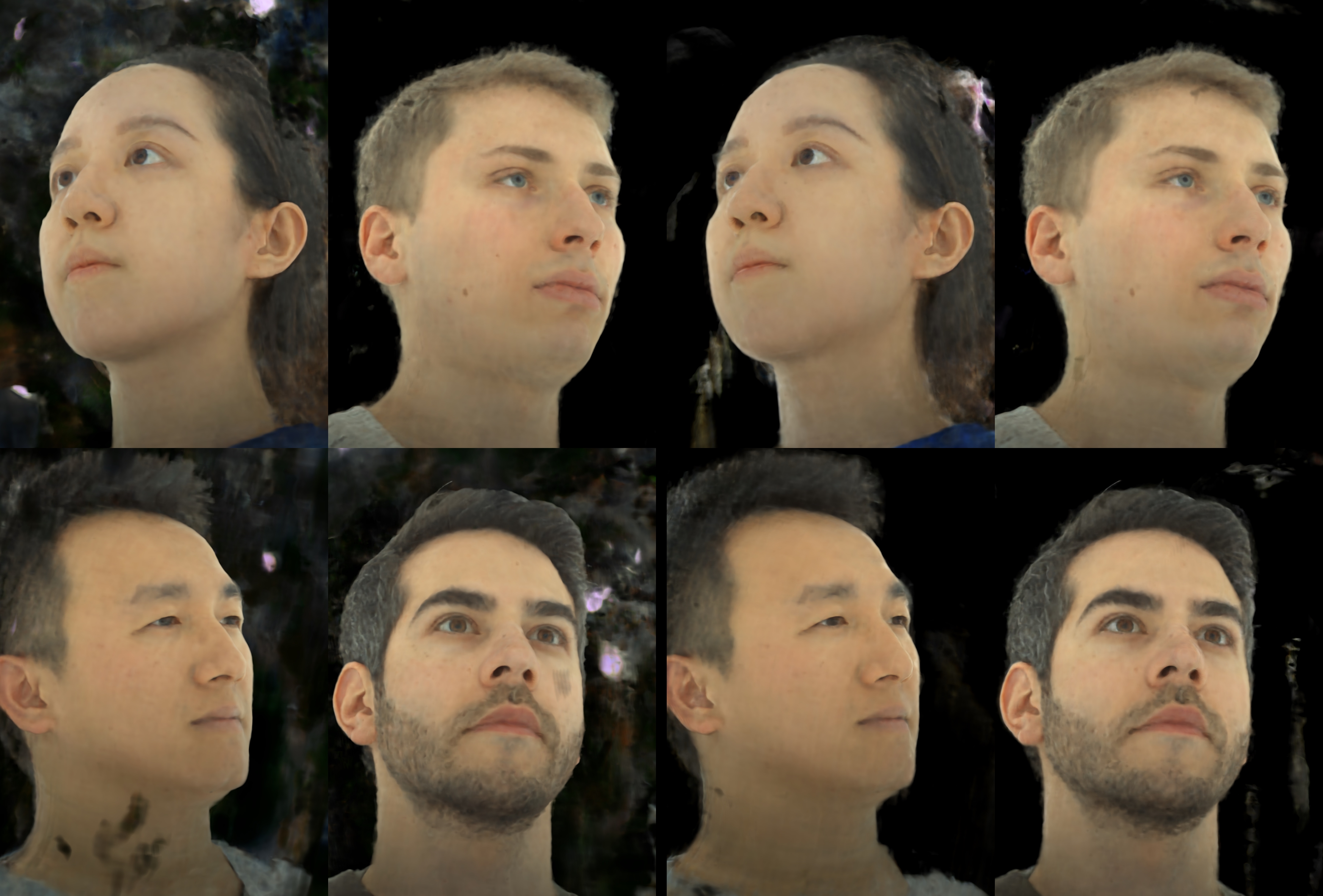}
\end{center}
   \caption{Examples of qualitative results from baseline and T-Face. Left: baseline. Right: T-Face.}
\label{fig:tface}
\end{figure}

\begin{figure}[h]
\begin{center}
   \includegraphics[width=0.98\linewidth]{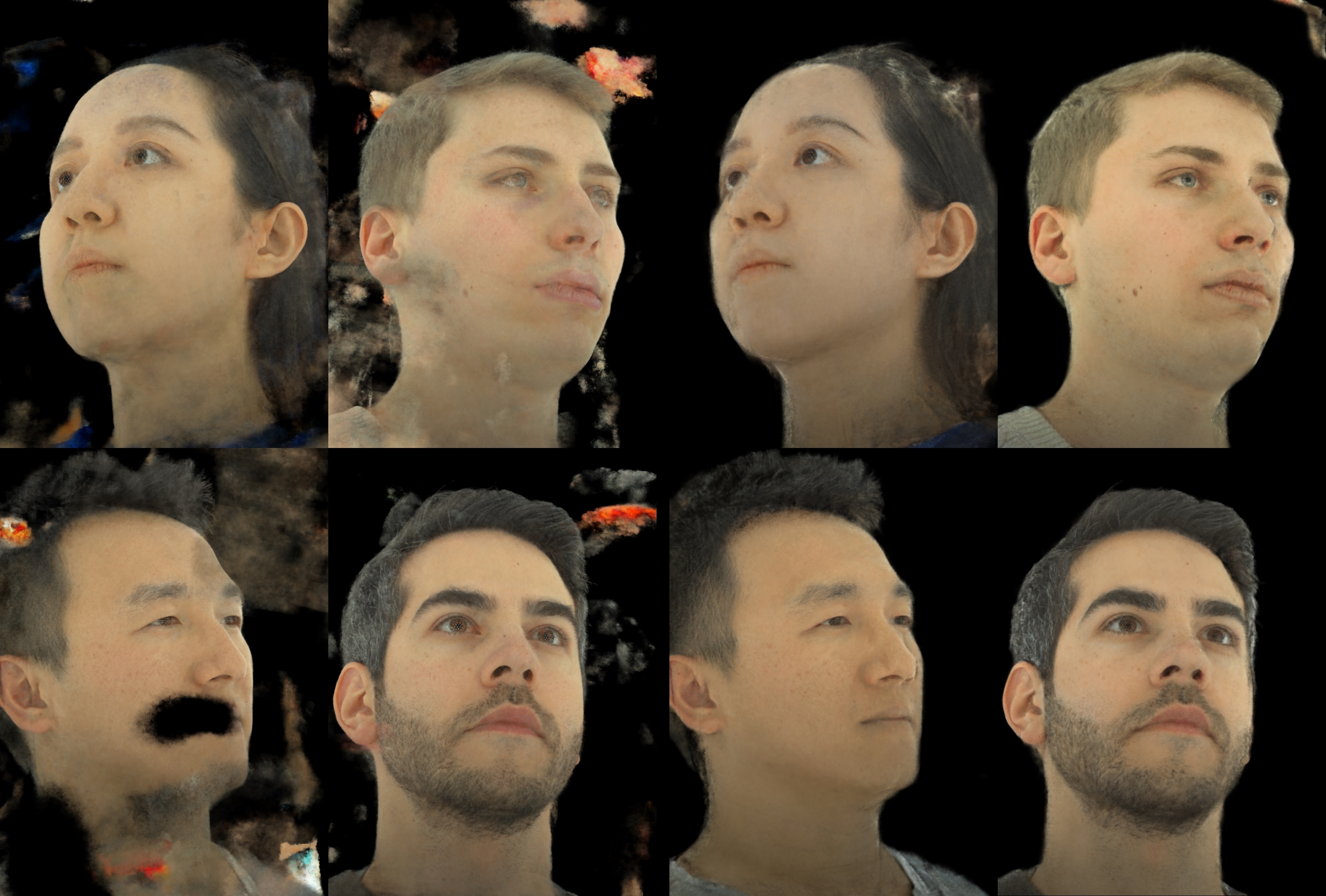}
\end{center}
   \caption{Examples of qualitative results from Instant-NeRF and I-Face. Left: Instant-NeRF. Right: I-Face.}
\label{fig:iface}
\end{figure}

\begin{figure}[h]
\begin{center}
   \includegraphics[width=0.98\linewidth]{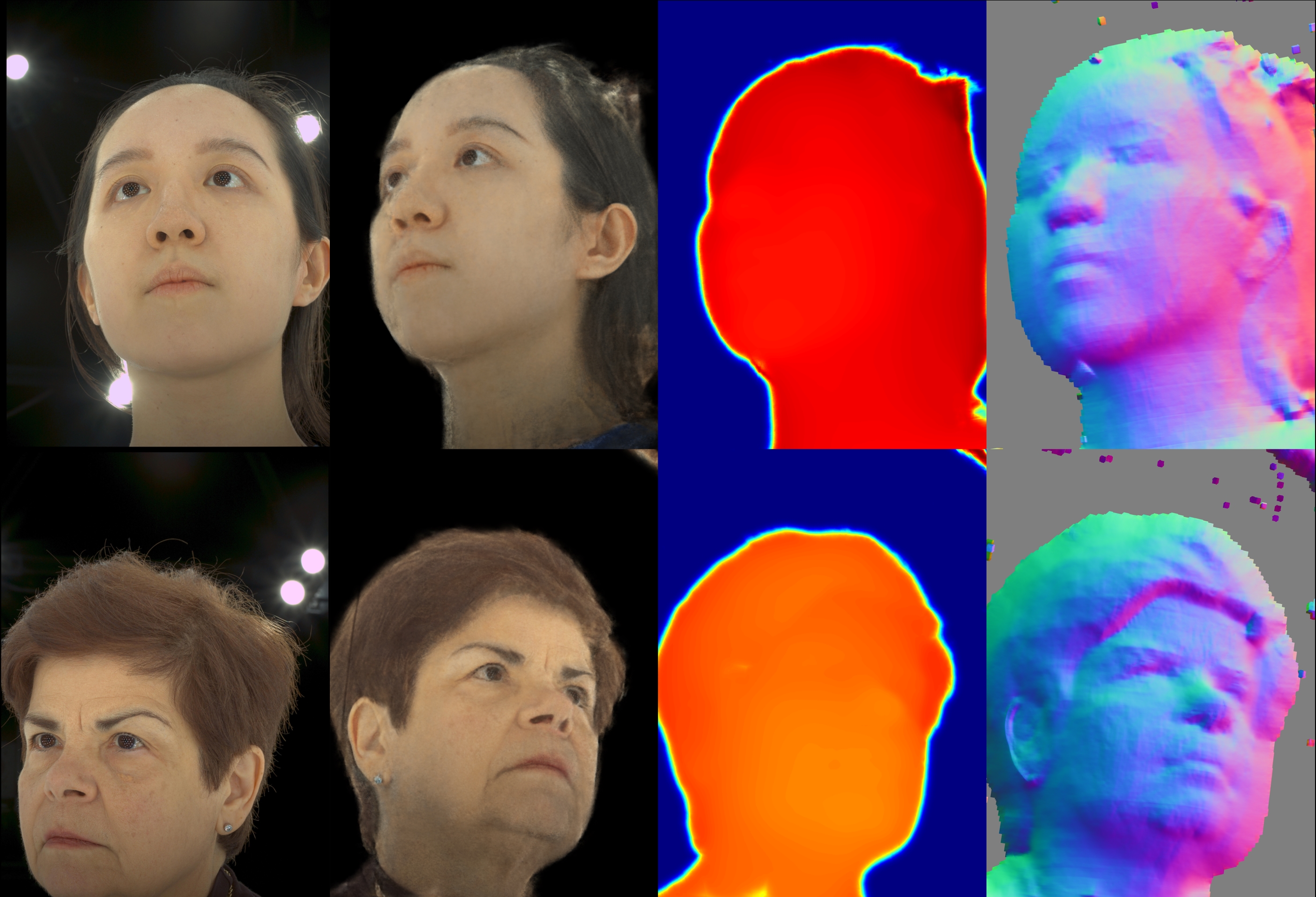}
\end{center}
   \caption{Examples of qualitative results from I-Face. From left to right: GT image (adjacent frames), rendered image, the predicted depth image, the predicted geometry normal.}
\label{fig:neus}
\end{figure}

\noindent{\textbf{Comparison with the baseline.}}
We evaluate our TI-Face in the ILSH dataset. As shown in~\Cref{table1}, the results obtained by emsembling baseline, T-Face and I-Face perform better than those rendered by either method alone. 
We also selected several examples to qualitatively compare these methods, as shown in~\Cref{fig:tface,,fig:iface}.
Interestingly, although I-Face appears to render clearer images, it does not perform as well as T-Face in actual evaluations. 
We speculate that the reason is that the results generated by I-Face have an overall deviation from the ground truth, as shown in the~\Cref{fig:neus}, the rendering results of I-Face have unexpected color differences and geometric inconsistency. 

\begin{table}
\begin{center}
\begin{tabular}{ccccc}
\hline
\multirow{2}{*}{Methods} & \multicolumn{2}{c}{Full Region} & \multicolumn{2}{c}{Masked Region} \\
\cmidrule(lr){2-3} \cmidrule(lr){4-5}
 & PSNR & SSIM & \textbf{PSNR} & SSIM \\
\hline
TensoRF~\cite{Chen2022tensorf} & 20.28 & \textbf{0.70} & 24.70 & 0.81 \\
T-Face(Ours) & 21.42 & 0.67 & 26.63 & 0.82 \\
I-Face(Ours) & 20.99 & 0.66 & 25.84 & 0.81 \\
TI-Face(Ours) & \textbf{21.66} & 0.68 & \textbf{27.02} & \textbf{0.83} \\
\hline
\end{tabular}
\end{center}
\caption{Results of baselines and our methods on the ILSH dataset in the challenge phase. TI-Face refers to the final fused results.}
\label{table1}
\end{table}

\begin{table}
\begin{center}
\begin{tabular}{ccccc}
\hline
\multirow{2}{*}{Settings} & \multicolumn{2}{c}{Masked Region} \\
\cmidrule(lr){2-3}
 & \textbf{PSNR} & SSIM \\
\hline
baseline & 24.03 & 0.83 \\
+ SAM mask & 25.24 & 0.83 \\
+ our mask & 25.39 & 0.84 \\
+ our mask + $\mathcal{L}_{mask}$(CE) & 25.52 & 0.84 \\
+ our mask + $\mathcal{L}_{mask}$(L2) & \textbf{26.77} & \textbf{0.84} \\
\hline
\end{tabular}
\end{center}
\caption{Abaltion study of T-Face on a subset (the first three) of ILSH dataset in the developing phase.}
\label{ablation}
\end{table}

\subsection{Ablation Study}
To demonstrate the effectiveness of the proposed improvement, we ablate our methods in the first three scenes of the ILSH dataset. 

\noindent{\textbf{Ablation of T-Face.}}
As shown in~\Cref{ablation}, the ablation experiments demonstrate the effectiveness of the proposed mask and corresponding loss function.
When adding the mask into baseline, we set the background region to black to eliminate artifacts. 
However, the results indicated that if the mask is not properly constrained, floating artifacts still persist, as shown in~\Cref{fig:baseline,,fig:mask}. 
Furthermore, the mask generated by our approach outperforms the mask generated directly by SAM, demonstrating the effectiveness of refining the mask generated by SAM. 
Furthermore, in the choice of the mask loss function, we surprisingly find that the use of L2 loss performs better than the use of cross entropy (CE) loss, resulting in a 1.25dB improvement in the PSNR metric in the masked region.

\begin{figure}[h]
\begin{center}
   \includegraphics[width=0.98\linewidth]{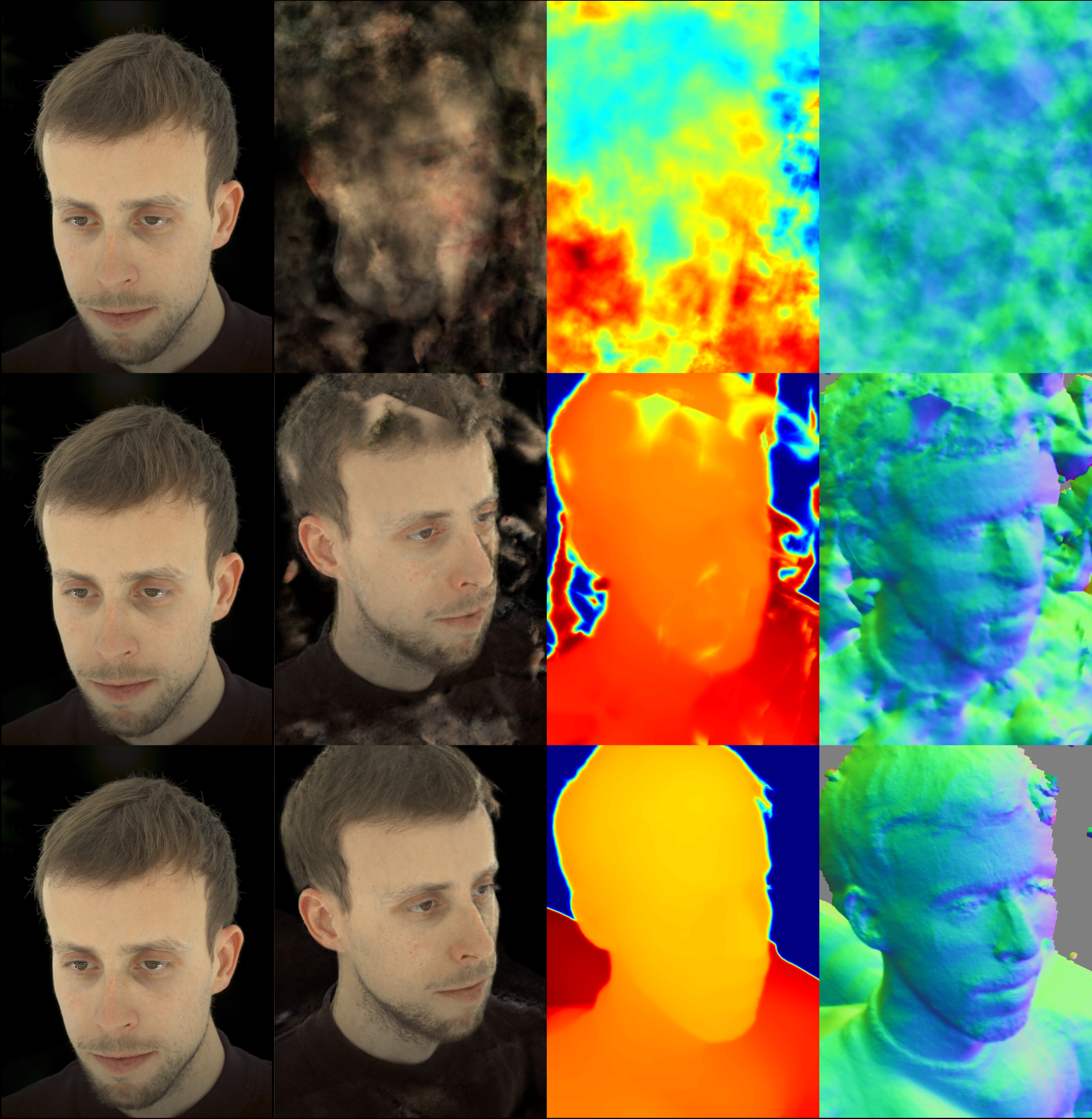}
\end{center}
\caption{Ablation study of I-Face. From left to right: GT image (adjacent frames), rendered image, the predicted depth image, the predicted geometry normal. From top to bottom: Instant-NeuS, Instant-NeuS+mask, I-Face(Instant-Neus+mask+$\mathcal{L}_{sparsity}$).}
\label{fig:neus_ablation}
\end{figure}

\noindent{\textbf{Ablation of I-Face.}}
The use of qualitative results to demonstrate the ablation experiment of I-Face is clearly more intuitive.
As shown in~\Cref{fig:neus_ablation}, the vanilla Instant-NeuS fails to reconstruct human head from unmasked sparse input views. 
After adding the generated mask, the proposed sparsity loss function $\mathcal{L}_{sparsity}$ further improves the quality of implicit surface reconstruction.
Although I-Face exhibits finer details visually, the overall reconstruction quality of I-Face is still not as high as that of T-Face. 
Our speculation is that this issue is attributed to overall color discrepancies caused by geometric inaccuracies..
Therefore, improving the rendering quality of implicit surfaces remains a huge challenge.

\section{Conclusion}

This report introduces the TI-Face model, designed to reconstruct human heads for novel view synthesis on a novel public dataset. The proposed model consists of two components: T-Face and I-Face, leveraging Tensorial Radiance Fields and Implicit Surfaces, respectively, to eliminate visual artifacts and enhance image rendering from new perspectives. The experiments demonstrate the effectiveness of the generated mask and the proposed loss function constraints, resulting in the first-place achievement at ICCV 2023 workshop VSCHH. Our proposed approach represents an exploration of explicitly modeling masks in the task of novel view synthesis. Investigating how to model masks for specific objects in more complex scenes and achieving reconstruction will be a valuable avenue for future research.

{\small
\bibliographystyle{ieee_fullname}
\bibliography{egbib}
}
\clearpage

\end{document}